\documentclass[10pt,journal,compsoc]{IEEEtran}

\ifCLASSOPTIONcompsoc
  \usepackage[nocompress]{cite}
\else
  \usepackage{cite}
\fi

\ifCLASSINFOpdf
\else
\fi

\usepackage[cmex10]{amsmath}
\usepackage{amssymb}
\usepackage[mathscr]{eucal}
\DeclareSymbolFont{rsfso}{U}{rsfso}{m}{n}
\DeclareSymbolFontAlphabet{\mathscr}{rsfso}
\usepackage[scr]{rsfso}
\usepackage{cite}
\usepackage{amsbsy}
\usepackage{esint}
\usepackage{tikz}
\usepackage{bbm}
\usepackage{graphbox}
\usepackage{graphicx}
\usepackage{caption}
\newsavebox{\imagebox}
\usepackage{subfigure}
\usepackage{makecell}
\usepackage{multirow}
\usepackage{url}
\usepackage{algorithm}
\usepackage{mathtools}
\usepackage[switch]{lineno}
\usepackage{booktabs}

                    {$\blacksquare$\vspace*{7pt}} 

\def\R{{\mathbb R}}

\def\N{{\mathbb N}}

\def\f{\frac}

\def\bi{{\mathbf i}}

\def\bi{\begin{itemize}} \def\ei{\end{itemize}}
\def\be{\begin{eqnarray*}}
\def\ee{\end{eqnarray*}}

\def\0{{\mathbf 0}}

\newcommand{\beq}{\begin{equation}}
\newcommand{\eeq}{\end{equation}}
\def\X{{\mathbf X}}

\def\XXint#1#2#3{{\setbox0=\hbox{$#1{#2#3}{\int}$ }
\vcenter{\hbox{$#2#3$ }}\kern-.55\wd0}}

\begin{document}

\title{A fully automated method for 3D individual tooth identification and segmentation in dental CBCT}

\author{Tae~Jun~Jang, Kang~Cheol~Kim, Hyun~Cheol~Cho, and Jin~Keun~Seo 
\IEEEcompsocitemizethanks{\IEEEcompsocthanksitem The authors are with School of Mathematics and Computing (Computational Science and Engineering), Yonsei University, Seoul, 03722.\protect\\ E-mail: kangcheol@yonsei.ac.kr (corresponding author)}}

\IEEEtitleabstractindextext{%
\begin{abstract}
Accurate and automatic segmentation of three-dimensional (3D) individual teeth from cone-beam computerized tomography (CBCT) images is a challenging problem because of the difficulty in separating an individual tooth from adjacent teeth and its surrounding alveolar bone. Thus, this paper proposes a fully automated method of identifying and segmenting 3D individual teeth from dental CBCT images. The proposed method addresses the aforementioned difficulty by developing a deep learning-based hierarchical multi-step model. First, it automatically generates upper and lower jaws panoramic images to overcome the computational complexity caused by high-dimensional data and the curse of dimensionality associated with limited training dataset. The obtained 2D panoramic images are then used to identify 2D individual teeth and capture loose- and tight- regions of interest (ROIs) of 3D individual teeth. Finally, accurate 3D individual tooth segmentation is achieved using both loose and tight ROIs. Experimental results showed that the proposed method achieved  an F1-score of 93.35\% for tooth identification and a Dice similarity coefficient of 94.79\% for individual 3D tooth segmentation. The results demonstrate that the proposed method provides an effective clinical and practical framework for digital dentistry.
\end{abstract}
\begin{IEEEkeywords}
Cone-beam computerized tomography, digital dentistry, tooth segmentation, tooth identification, deep learning
\end{IEEEkeywords}}

\maketitle

\IEEEdisplaynontitleabstractindextext

\IEEEpeerreviewmaketitle

\ifCLASSOPTIONcompsoc
\IEEEraisesectionheading{\section{Introduction}\label{sec:introduction}}
\else
\section{Introduction}
\label{sec:introduction}
\fi


\IEEEPARstart{D}{igital} dentistry is evolving rapidly along with the rapid innovation of artificial intelligence and the development of cone-beam computerized tomography (CBCT), intra-oral and facial scanners, and dental three-dimensional (3D) printing. Digital dentistry enhances a dentist's efficiency and improves the accuracies of orthodontic diagnoses, treatment planning, and surgical guides. A fundamental component of digital dentistry is the 3D segmentation of teeth, jaws, and skulls from CBCT images. Moreover, accurate digital models of individual tooth geometry and jaws facilitate the simulation of prosthetic evaluation, cephalometric analysis, computer-aided digital implant planning, and bite irregularity prediction.
	
Automatic and accurate 3D individual tooth segmentation from CBCT images is a difficult task for the following reasons: (i) similar intensities between teeth roots and their surrounding alveolar bone; (ii) attached boundary between adjacent teeth in the crown parts.

Over the last decade, there have been several attempts to develop 3D tooth segmentation methods, most of which are based on level set methods \cite{gao2010, gan2015, yau2014}. Unfortunately, level set-based methods have fundamental limitations in achieving fully automated segmentation. This difficulty arises from the dependence of such methods on the initialization of level set, and the automatic initialization is hindered by the complex image structure associated with adjacent teeth, the jaw, the alveolar bone, etc. Hence, user intervention through manual initialization is inevitable in this approach.

Recently, deep learning methods have been applied in 3D tooth segmentation. Lee \textit{et al.} \cite{lee2020} and Rao \textit{et al.} \cite{rao2020} used a fully convolutional network (FCN) \cite{long2015} for whole tooth segmentation instead of individual tooth segmentation. Cui \textit{et al.} \cite{cui2019} proposed a deep learning framework for individual tooth segmentation and identification using Mask R-CNN \cite{he2017}. The limitation of these deep learning methods is patch-based approach to handle high-dimension inputs (\textit{e.g.}, $800\times800\times400$ voxels in 3D CBCT image) and a limited amount of labeled samples. It is necessary to use both local and global information to achieve accurate segmentation with individual tooth identification. Thus, the drawback of this patch-based approach is its inability to reflect contextual (global) information, since each output of a convolutional network only depends on the corresponding patch. 

Automatic individual tooth identification is also a difficult task. Recently, several individual tooth identification attempts \cite{miki2017, tuzoff2019} have been made using convolutional networks. However, these approaches suffer from misclassification errors caused by adjacent teeth similarities.

\begin{figure*}[h]
\centering
\includegraphics[width=0.925\textwidth]{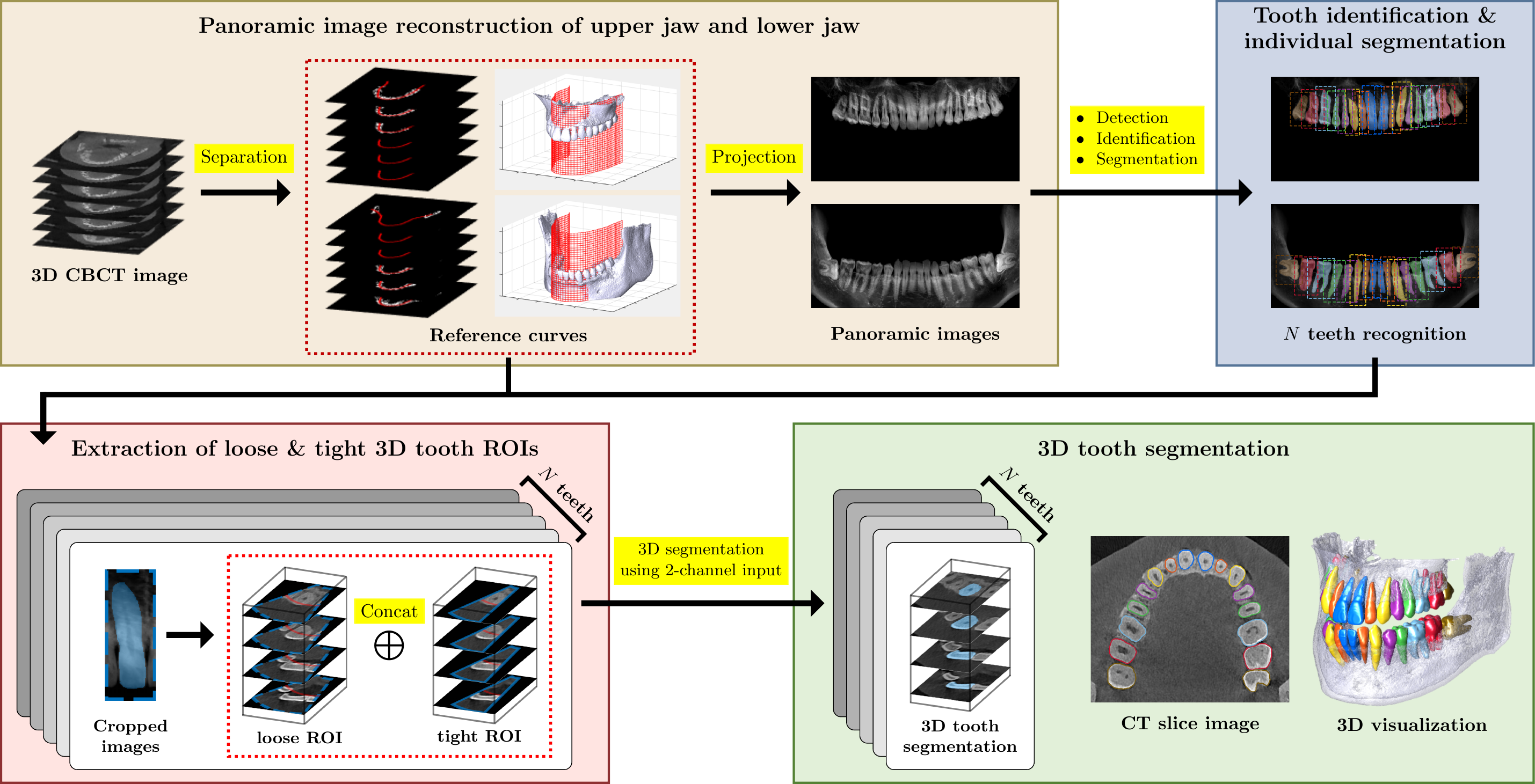}
\caption{Schematic diagram of the proposed method, which consists of four steps: 1) Panoramic image reconstruction of the upper and lower jaws from a 3D CBCT image; 2) tooth identification and 2D segmentation of individual teeth in the panoramic images; 3) extraction of loose and tight 3D tooth ROIs using the detected bounding boxes and segmented tooth regions; and 4) 3D segmentation for individual teeth from the 3D tooth ROIs.}
\label{fig:Framework}
\end{figure*}

Existing 3D tooth segmentation methods may not be effective for  CBCT images that are severely corrupted by metal artifacts. In a clinical dental CBCT environment (\textit{e.g.}, low dose radiation exposure), metal artifacts become common as the number of aged patients with metallic prosthesis increases. Hence, to be practical, it would be desirable to develop a method that works well even in images degraded by metal artifacts.

This paper aimed to address these limitations by developing a hierarchical multi-step deep learning model. The proposed method is summarized as follows. The first step is to circumvent the high-dimensionality problem associated with CT images. This step automatically generates panoramic images of the upper and lower jaws from CT images where its size is smaller than the original CT image. The panoramic images of the upper and lower jaws are separated to reduce overlaps between adjacent teeth. Notably, panoramic images generated from CBCT images are not significantly affected by metal-related artifacts. We utilize these panoramic images to accurately perform 2D tooth detection, identification, and segmentation. The second step is to identify individual teeth by numbers according to F\'ed\'eration Dentaire Internationale (FDI) dental notation. We develop a tooth detection method to localize bounding boxes that enclose each tooth and classifies them into four types (incisors, canines, premolars, and molars) according to tooth morphology. This method solve misclassification problems caused by similar adjacent teeth. The individual teeth are then identified using the results of tooth detection. Additionally, we perform 2D segmentation for individual teeth. The third step extracts loose and tight 3D tooth regions of interest (ROIs) from the detected boxes and segmented tooth regions for accurate 3D individual tooth segmentation in the final step. Tight ROIs improve the segmentation accuracy. A schematic diagram of our method is described in Fig. \ref{fig:Framework}.

\section{Method}
\label{sec:method}
\begin{figure*}[!h]
\centering
\includegraphics[width=0.925\textwidth]{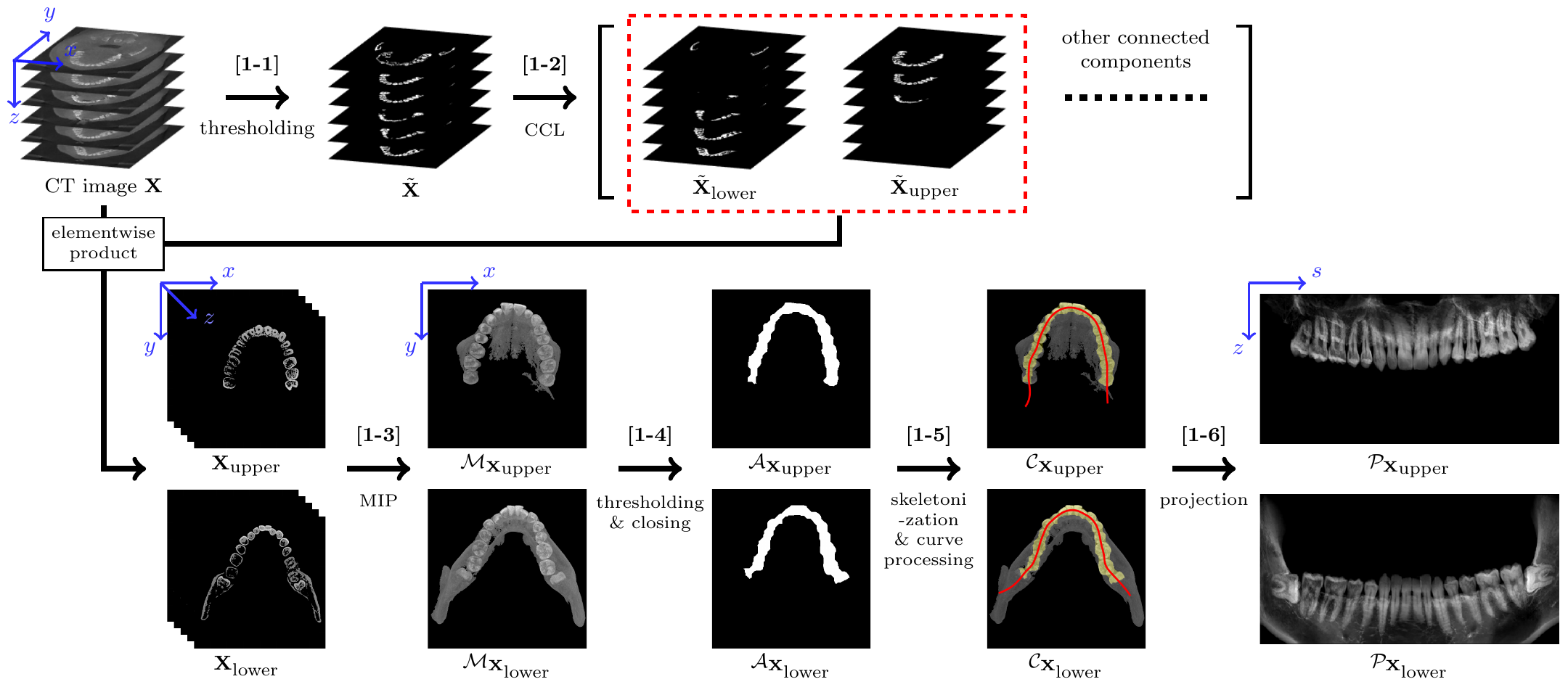}					 					  		
\caption{Workflow of Step 1. This shows reconstruction process of upper jaw panoramic image $P_{\bf{X}_{\mbox{\scriptsize upper}}}$ and lower jaw panoramic images $P_{\bf{X}_{\mbox{\scriptsize lower}}}$ from a 3D CT image $\bf X$.}
\label{fig:step1}
\end{figure*}

Let $\bf{X}$ denote a 3D CT image with voxel grid $\Omega:=\{ (x,y,z) \in \N^{3}: 1 \leq x \leq N_{x}, 1 \leq y \leq N_{y}, 1 \leq z \leq N_{z} \}$, where $N_{x}$, $N_{y}$ and $N_{z}$ are the voxel sizes in directions $x$ (sagittal axis), $y$ (frontal axis) and $z$ (longitudinal axis), respectively. The value ${\bf{X}}(x,y,z)$ at the voxel position $(x,y,z)$ is represented as the attenuation coefficient.

\subsection{Step 1: Panoramic image reconstruction of the upper and lower jaws from a 3D CBCT image}

This step describes the automatic reconstruction of the upper and lower jaw panoramic images from a 3D CBCT image $\bf X$. Fig. \ref{fig:step1} illustrates the workflow.

[Step 1-1] To obtain a binary bone image $\tilde{\X}$, a 3D CT image $\X$ is segmented into three classes (air, soft tissues, and bones) using multi-level version of Otsu's method \cite{Otsu1979}. The threshold values $T_0$ and $T_1$ for the histogram $h(t)$ corresponding to $\X$ are determined by

	\begin{equation}
	\label{eq:Otsu}
	\begin{split}
	\{T_0, T_1\}~ = ~\underset{t_0,t_1} {\mbox{argmax}} \left[ \mbox{\small $\left( \f {\underset{0\le i<t_0}{\sum}ip(i)}{\underset{0\le i<t_0}{\sum}p(i)} - {\underset{0\le i<t_0}{\sum}p(i)} \right)^2$} \right.  ~~~~~~~~~~~~~~&\\
	+ \mbox{\small $\left( \f {\underset{t_0\le i<t_1}{\sum}ip(i)}{\underset{t_0\le i<t_1}{\sum}p(i)} - {\underset{t_0 \le i<t_1}{\sum}p(i)} \right)^2$} + \left. \mbox{\small $\left( \f {\underset{i \ge t_1}{\sum}ip(i)}{\underset{i \ge t_1}{\sum}p(i)} - {\underset{i \ge t_1}{\sum}p(i)} \right)^2$}  \right],&
	\end{split}
	\end{equation}
where $p(t)=h(t)/ \underset{ i}{\sum} h(i)$. The binary image $\tilde{\X}(x,y,z)$ is 1 if $\X(x,y,z) \geq T_1$, and 0 otherwise. As this value $T_1$ corresponds to an interface between soft tissues and bones, $\tilde{\X}$ can be viewed as a rough segmentation of upper and lower jaws.

[Step 1-2] Given the binary image $\tilde{\X}$, the connected-component labeling (CCL) \cite{Samet1988} is used to extract the upper jaw part $(\tilde{\X}_{\mbox{\tiny upper}})$ and the lower jaw part $(\tilde{\X}_{\mbox{\tiny lower}})$. The CCL method generates all the connected components in a binary image. The lower jaw part is the largest connected component and the upper jaw part is the second largest connected component.

[Step 1-3] To create a 2D image $\mathcal{M}_{\bold{X}_{\mbox{\tiny upper}}}$ displaying the upper dental arch, we apply maximum intensity projection (MIP) in the $z$ direction to the grayscale upper jaw image ($\X_{\mbox{\tiny upper}}=\X \odot \tilde{\X}_{\mbox{\tiny upper}}$, where $\odot$ is an elementwise product), as follows:
	\begin{equation}
	\mathcal{M}_{\X_{\mbox{\tiny upper}}}(x,y)=\underset{z}{\max}~ \X_{\mbox{\tiny upper}}(x,y,z).
	\end{equation}
	Similarly, we obtain $\X_{\mbox{\tiny lower}}$ and $\mathcal{M}_{\X_{\mbox{\tiny lower}}}$ for the lower jaw.

[Step 1-4] Next, binary dental arch regions $\mathcal{A}_{\X_{\mbox{\tiny upper}}}$ and $\mathcal{A}_{\X_{\mbox{\tiny lower}}}$ is obtained by applying the Otsu's method \cite{Otsu1979} and the morphological closing \cite{Haralick1987} to the MIP images $\mathcal{M}_{\X_{\mbox{\tiny upper}}}$ and $\mathcal{M}_{\X_{\mbox{\tiny lower}}}$, respectively. Here, Otsu thresholding is adopted to get rough dental arch regions and the morphological closing is used to smoothen out the rough regions.

[Step 1-5] Given the upper dental arch region $\mathcal{A}_{\X_{\mbox{\tiny upper}}}$, we employ the morphological skeletonization \cite{Lee1994} to extract a medial axis of the dental arch region. Cubic spline curve fitting, interpolation and extrapolation techniques are then applied to the medial axis, to obtain a smooth reference curve $\mathcal{C}_{\X_{\mbox{\tiny upper}}}$ passing through the dental arch region completely. The reference curve can be expressed as
	\begin{equation}	
	\mathcal{C}_{\X_{\mbox{\tiny upper}}} = \{ \bold{r}(s)=(x(s),y(s)): s \in {1,2,\cdots,N_{s}} \},
	\label{eq:curve}
	\end{equation}
	where $N_s$ is the number of curve points. Similarly, we can obtain $\mathcal{C}_{\X_{\mbox{\tiny lower}}}$ from the lower dental arch region $\mathcal{A}_{\X_{\mbox{\tiny lower}}}$.

[Step 1-6] An upper jaw panoramic image is given by
	\begin{equation}
	P_{\X_{\mbox{\tiny upper}}}(s,z) = \int^{\alpha}_{-\alpha} \X_{\mbox{\tiny upper}} \left( \bold{r}(s)+t\bold{n}(s),z \right)\,dt,
	\label{eq:pan}
	\end{equation}
	where $s$ is the parameter in \eqref{eq:curve}, $\bold{r}(s) \in \mathcal{C}_{\X_{\mbox{\tiny upper}}}$, and $\bold{n}(s)$ is the unit normal vector at $\bold{r}(s)$. Similarly, we obtain the lower panoramic image $P_{\X_{\mbox{\tiny lower}}}$. For notational simplicity, we refer to both $P_{\X_{\mbox{\tiny upper}}}$ and $P_{\X_{\mbox{\tiny lower}}}$ as $P$.

\subsection{Step 2: Bounding box detection, identification, and 2D segmentation of individual teeth in the reconstructed panoramic images}

This step aims to identify and segment individual teeth in the reconstructed panoramic images. To achieve the goal, we first perform individual tooth detection. Here, the teeth are classified as incisor (class 1), canine (class 2), premolar (class 3), and molar (class 4). 

\begin{figure}[h!]
  \centering
  \includegraphics[width=0.4\textwidth]{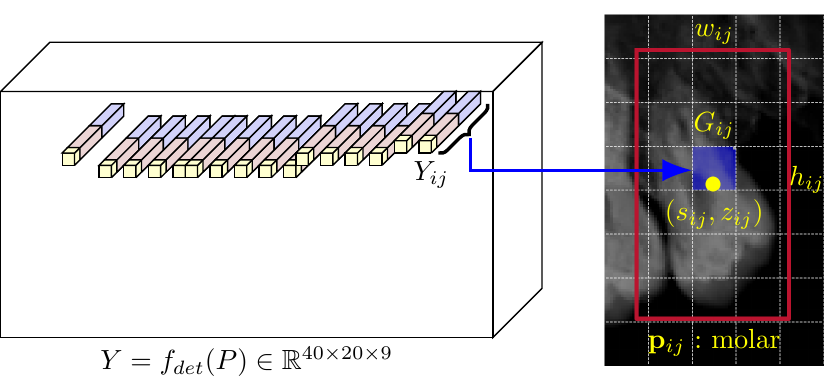}
  \caption{Concept of Step 2-1. A detection map $f_{det}$ predicts $Y_{ij}$ for each $G_{ij}$.}\label{fig:step2-1}
\end{figure}	

[Step 2-1] To detect individual teeth in a panoramic image, we develop a deep learning method inspired by one-stage object detection \cite{redmon2016, liu2016}. Given a panoramic image $P$ with size $N_s \times N_z$ ($e.g.,~N_s\times N_z=640\times320$), a uniform grid is created. Each grid cell $G_{ij}$ has a size of $g \times g$ ($e.g.,~g=16$).
	Then we learn a tooth detection map $f_{det}:P \mapsto Y$ that is given by
	\begin{equation}
	f_{det}(P)=
	\left(
	\begin{array}{cccc}
	Y_{1,1} & Y_{2,1} & \cdots & Y_{\frac{N_s}{g},1} \\
	Y_{1,2} & Y_{2,2} & \cdots & Y_{\frac{N_s}{g},2}\\
	\vdots &  & \ddots & \vdots \\
	Y_{1,\frac{N_z}{g}} & \cdots & \cdots & Y_{\frac{N_s}{g},\frac{N_z}{g}}
	\end{array}
	\right),
	\end{equation}
	where $Y_{ij}=(c_{ij},\bold{b}_{ij},\bold{p}_{ij})$ predicting a confidence score $c_{ij}$, a bounding box component $\bold{b}_{ij}$, and a class probability $\bold{p}_{ij}$ in $G_{ij}$, as illustrated in Fig. \ref{fig:step2-1}. Here, a confidence score $c_{ij}\in[0,1]$ represents the existence of the tooth center in $G_{ij}$. A bounding box component \cite{girshick2014} is denoted by
	\begin{equation}
	\label{eq:box}
	\bold{b}_{ij}=\left(\frac{s_{ij}}{g}-(i-1),\frac{z_{ij}}{g}-(j-1),\log{\frac{w_{ij}}{a_w}},\log{\frac{h_{ij}}{a_h}}\right),
	\end{equation}
	where $(s_{ij},z_{ij})$ is the center of the bounding box in $G_{ij}$, $(w_{ij},h_{ij})$ indicates its width and height, and $(a_w,a_h)$ are the width and height of an anchor box. For a tooth in the bounding box corresponding to $\bold{b}_{ij}$, we estimate a class probability
	\begin{equation}
	\label{eq:prob}
	\bold{p}_{ij}=(p_{ij,1},p_{ij,2},p_{ij,3},p_{ij,4}),
	\end{equation}
	where $p_{ij,k}$ represents the probability of being tooth class $k$.

To find exact bounding boxes among the predicted boxes for all $G_{ij}$, we remove the boxes with scores $e_{ij}=c_{ij}*(\underset{k}{\max}~p_{ij,k})$ less than 0.5. Several bounding boxes with high scores may appear near the center of a tooth. We adopt the non-maximum suppression (NMS) technique to filter out bounding boxes that highly overlap high-scoring boxes.
	
Using a labeled training dataset $\{(P^{(n)},Y^{*(n)})\}_{n=1}^{N}$ where $Y^*$ is ground-truth labeling, $f_{det}$ is learned by minimizing the loss between the output $Y=f_{det}(P)$ and the ground-truth $Y^*$ as follows:
	\begin{equation}
	\begin{split}
		\mathcal{L}_{det}=\underset{n=1}{\overset{N}{\sum}} \left[ \underset{(i,j) \in \Omega_{1}^{(n)}}{\sum} (1-c_{ij}^{(n)})^2 + \lambda_1 \underset{(i,j) \in \Omega_{0}^{(n)}}{\sum} (0-c_{ij}^{(n)})^2 \right.\\
		\left. + \lambda_2\underset{(i,j) \in \Omega_{1}^{(n)}}{\sum}|\bold{b}_{ij}^{*(n)}-\bold{b}_{ij}^{(n)}|^2 + \underset{(i,j) \in \Omega_{1}^{(n)}}{\sum}\mbox{CE} (\bold{p}^{*(n)}_{ij},{\bold{p}_{ij}^{(n)}}) \right],
	\end{split}
	\label{eq:lossDet}
	\end{equation}
	where $\Omega_{1}^{(n)}=\{(i,j)|c^*_{ij}=1\}$, $\Omega_{0}^{(n)}=\{(i,j)|c^{*}_{ij}=0\}$, and $\mbox{CE}$ is the cross-entropy. This multi-task loss $\mathcal{L}_{det}$ represents the prediction errors where objects exist ($c^*_{ij}=1$) and no objects exist ($c^*_{ij}=0$). Since there is no object in most grid cells, the confidence score tends to be predicted as zero \cite{redmon2016}. The constants $\lambda_1=0.1$ and $\lambda_2=5$ are used to eliminate the imbalance.

\begin{figure}[h]
\centering
\includegraphics[width=0.36\textwidth]{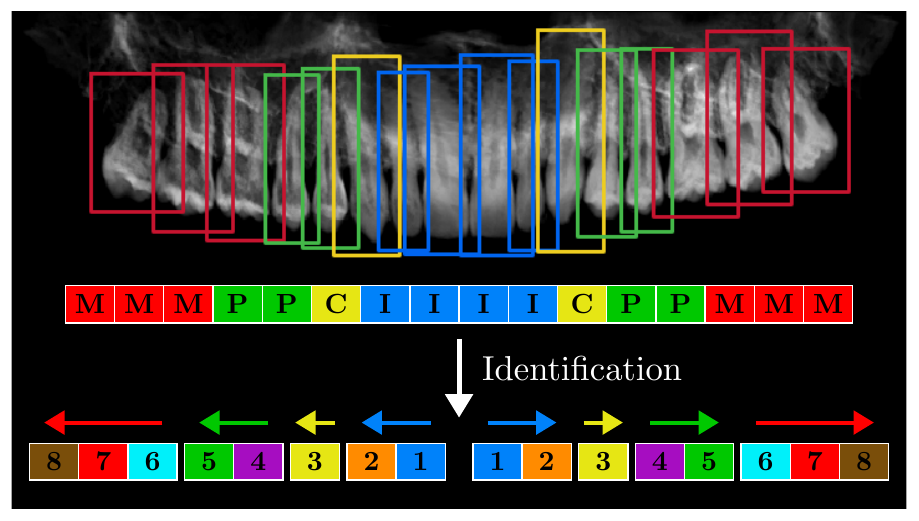}
\caption{Tooth identification process using the classification results in Step 2-1. The capital letters represent the first letters of the tooth type and the numbers are tooth codes.}
\label{fig:step2-2}
\end{figure}

[Step 2-2] For each tooth in the detected bounding box, a number is assigned to identify the unique tooth according to the FDI system. For convenience, we first explain how the numbers are assigned to teeth in the upper jaw. As illustrated in Fig. \ref{fig:step2-2}, the detected bounding boxes are listed in ascending order of $s$ coordinates of the box center. The upper right and left quadrants are divided from the middle of four sequential incisor boxes. For the two right incisors and the two left incisors, number 1 and 2 are assigned from the inside to the outside, respectively. Number 3 is assigned to the canines since there is only one in each quadrant. On each side, premolars are assigned numbers 4 and 5 from the inside to the outside. Likewise, molars are assigned numbers 6, 7 and 8 (if a wisdom tooth exists).

\begin{figure*}[h!]
\centering
\includegraphics[width=0.87\textwidth]{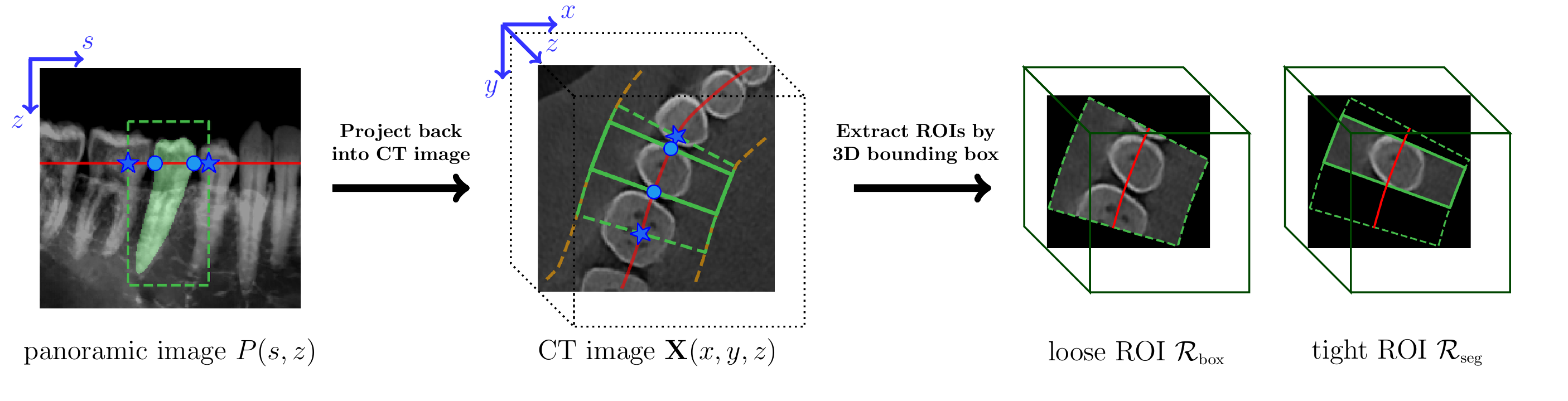}
\caption{Extraction of loose and tight 3D tooth ROIs from a detected bounding box and a segmented tooth region.}

\label{fig:step3}
\end{figure*}

[Step 2-3]	The proposed 2D tooth segmentation method uses a U-shaped FCN \cite{ronneberger2015} with taking advantage of bounding box knowledges obtained from Step 2-1. Let $S\in\R^{N_s \times N_z}$ denote the segmentation image for a tooth corresponding to a detected bounding box in $P$. We construct a training dataset $\{I^{(n)}_{\mbox{\scriptsize roi}},Y^{(n)}_{\mbox{\scriptsize roi}}\}_{n=1}^{N}$ for individual teeth segmentation, where $I^{(n)}_{\mbox{\scriptsize roi}}$ and $Y^{(n)}_{\mbox{\scriptsize roi}}$ are tooth images of $P$ and $S$ cropped by the bounding boxes. A segmentation map $f_{seg}:I_{\mbox{\scriptsize roi}} \mapsto Y_{\mbox{\scriptsize roi}}$ is learned using a U-shaped network and minimizing the following loss:
	\begin{equation}
	\mathcal{L}_{seg}=\frac{1}{N}\sum_{n=1}^{N} \left[ -\frac{1}{M} \sum _{\bold{x}} Y^{(n)}_{\mbox{\scriptsize roi}}(\bold{x}) \log\left[f_{seg}\left(I^{(n)}_{\mbox{\scriptsize roi}}\right)(\bold{x})\right] \right],
	\label{eq:lossSeg}
	\end{equation}
	where $\bold{x}$ is a pixel position and $M$ is the number of pixels of $Y_{\mbox{\scriptsize roi}}$.

\subsection{Step 3: Extraction of loose and tight 3D tooth ROIs using the detected bounding boxes and segmented tooth regions}
	
In this step, 3D tooth ROIs are obtained using the results of the previous steps. As described in Fig. \ref{fig:step3}, a bounding box containing one tooth is projected back into the 3D CBCT image using \eqref{eq:curve} and \eqref{eq:pan}. A loose ROI domain of the target tooth is then given by
	\begin{equation}
	D_{\mbox{\scriptsize box}} = \{(\bold{r}(s)+t\bold{n}(s),z): -\alpha \leq t \leq \alpha,~(s,z)\in B_{\mbox{\scriptsize box}}\},
	\end{equation}
	where $B_{\mbox{\scriptsize box}}$ is the set of pixel positions in the bounding box. Similarly, a tight ROI domain $D_{\mbox{\scriptsize seg}}$ is determined by $B_{\mbox{\scriptsize seg}}$, which is the set of pixel positions in the 2D tooth segmented region in the box.

The loose 3D tooth ROI $\mathcal{R}_{\mbox{\scriptsize box}}$ is obtained by changing the voxel values outside $D_{\mbox{\scriptsize box}}$ to 0, and extracting content in a 3D bounding box that fits closely around $D_{\mbox{\scriptsize box}}$, as shown in Fig. \ref{fig:step3}. Similarly, we obtain the tight 3D tooth ROI $\mathcal{R}_{\mbox{\scriptsize seg}}$ by processing $D_{\mbox{\scriptsize seg}}$ instead of $D_{\mbox{\scriptsize box}}$, and using the same 3D bounding box as above.

\subsection{Step 4: 3D segmentation for individual teeth from the 3D tooth ROIs}

In this final step, 3D individual tooth segmentation is performed by applying the loose ROI ($\mathcal{R}_{\mbox{\scriptsize box}}$) and tight ROI ($\mathcal{R}_{\mbox{\scriptsize seg}}$) to the 3D version of a U-shaped FCN \cite{ronneberger2015}. The tight ROI is crucial for improving the segmentation accuracy at the attached boundaries between a target tooth and its neighboring teeth.

The input of the network is $I_{\mbox{\scriptsize roi3}}=\mathcal{R}_{\mbox{\scriptsize box}} \oplus \mathcal{R}_{\mbox{\scriptsize seg}}$, which represents the concatenating vector of two ROIs. Let $Y_{\mbox{\scriptsize roi3}}$ denote a binary vector representing 3D tooth segmentation corresponding to $I_{\mbox{\scriptsize roi3}}$. Using a training dataset $\{I^{(n)}_{\mbox{\scriptsize roi3}},Y^{(n)}_{\mbox{\scriptsize roi3}}\}^{N}_{n=1}$, we learn a 3D segmentation map $f_{seg3}:I_{\mbox{\scriptsize roi3}} \mapsto Y_{\mbox{\scriptsize roi3}}$ by minimizing the following loss:
	\begin{equation}
	\mathcal{L}_{seg3}=\frac{1}{N}\sum_{n=1}^{N} \left[ -\frac{1}{V} \sum _{\bold{v}} Y_{\mbox{\scriptsize roi3}}^{(n)}(\bold{v}) \log\left[f_{seg3}\left(I_{\mbox{\scriptsize roi3}}^{(n)}\right)(\bold{v})\right] \right],
	\label{eq:lossSeg3}
	\end{equation}
	where $\bold{v}$ is a voxel position and $V$ is the number of voxels of $Y_{\mbox{\scriptsize roi3}}$.

\section{Experiments and Results}
\label{sec:exp}

\subsection{Dataset and implementation details}

Experiments were conducted on 3D CT images produced by a dental CBCT with a circular trajectory (DENTRI-X; HDXWILL, Seoul, South Korea) using tube voltage 90kVp and tube current 10mA. All available datasets were formatted in the Digital Imaging and Communications in Medicine (DICOM) standard. The size of the CT image was $800\times800\times400$, and its pixel spacing and slice thickness were both 0.2mm. During scanning, a bite block was used to prevent contact between the upper and lower teeth.

We received 97 dental 3D CBCT images from HDXWILL. Using these data, we generated 194 upper and lower jaws panoramic images in Step 1. We also received labeled data consisting of 97 2D individual tooth segmentation, bounding box components, and tooth codes, as well as 11 3D individual tooth segmentation. The labeling was performed by experts in HDXWILL.

Panoramic images were generated by \eqref{eq:pan} using a tomographic reconstruction software (TIGRE) \cite{biguri2016}. The size of all panoramic images was fixed at $640\times320$. The width of the panoramic images was determined by 640 reference curve points in Step 1-5. Those points were obtained by interpolating 500 points on the smooth curve and extrapolating 70 points at each end of the curve. To completely cover the teeth at both ends, we extrapolated 70 points (approximately 13.3mm) taking into account the average size of the molars. The height size 320 was determined by removing 80 CT slices that do not contain teeth from the bottom.

For the 2D detection and segmentation in Steps 2-1 and 2-3, 66 CBCT dataset were used for training and 31 dataset for testing. Because two panoramic images (upper and lower parts) were obtained from each CBCT image through Step 1, we used 132 labeled training data and 62 test data. Meanwhile, for the 3D segmentation in Step 4, 7 CBCT dataset were used for training and 4 dataset for testing. Since each patient has approximately 28 to 32 teeth, each CBCT image can provide approximately 28 to 32 training data for individual tooth segmentation. To be precise, we used 216 training data and 112 test data for the 3D segmentation in Step 4. In Steps 2-3 and 4, 2D tooth images and 3D loose and tight ROIs were resized to $128\times128$ and $128\times128\times128$, respectively.

The proposed neural networks were trained by minimizing losses in \eqref{eq:lossDet}, \eqref{eq:lossSeg}, and \eqref{eq:lossSeg3} using Adam optimizer \cite{kingma2014} from PyTorch \cite{paszke2019}. We used five-fold cross validation on the training dataset. Batch sizes of 8, 32, and 4 were set for each training in Steps 2-1, 2-3, and 4, respectively. We analyzed the runtime of the proposed convolutional networks on a desktop with a GeForce RTX 2070 GPU. The processing time of the proposed three networks in Steps 2-1, 2-3, and 4 were 3.62, 5.98, and 6.82 ms/batch, respectively.

\subsection{Evaluation and Result of the proposed method}

\subsubsection{Bounding box detection}

\begin{figure}[!h]
  \centering
  \includegraphics[width=0.39\textwidth]{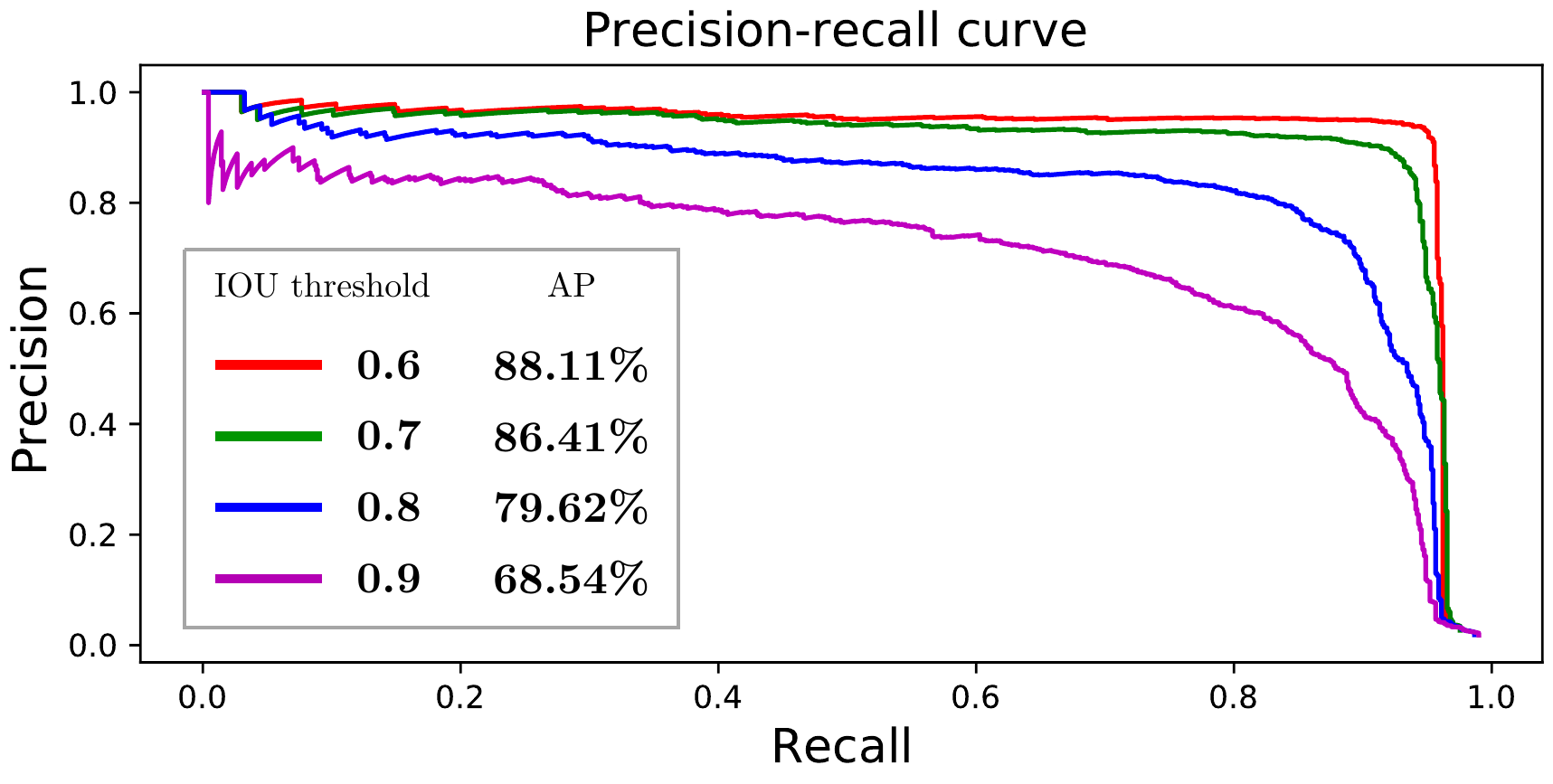}
  \caption{Tooth detection results. A PR curve represents the change in the precision as the recall increases for a fixed IOU threshold value, which is used for NMS.}\label{fig:pr_curve}
\end{figure}

For a quantitative evaluation of the bounding box detection, we provide precision-recall (PR) curves \cite{everingham2010} and their average precision (AP) \cite{everingham2010}, as shown in Fig. \ref{fig:pr_curve}. When the intersection over union (IOU) threshold value was 0.6, according to the PR curve, the precision tends to stay high as the recall increases, and the AP was $88.11\%$.

\subsubsection{Individual tooth identification}

\begin{figure}[!h]
	\centering
	\subfigure[]{\includegraphics[width=0.24\textwidth]{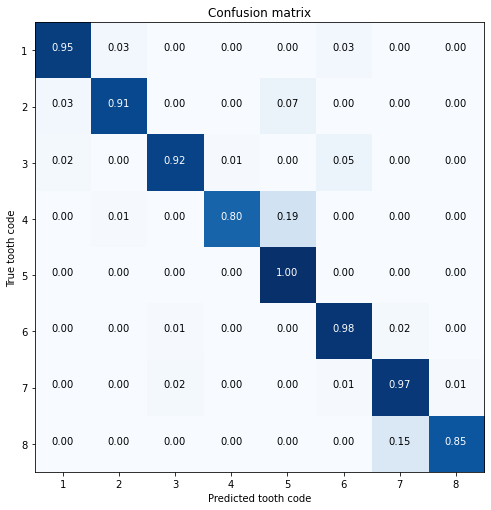}}~
	\subfigure[]{\includegraphics[width=0.24\textwidth]{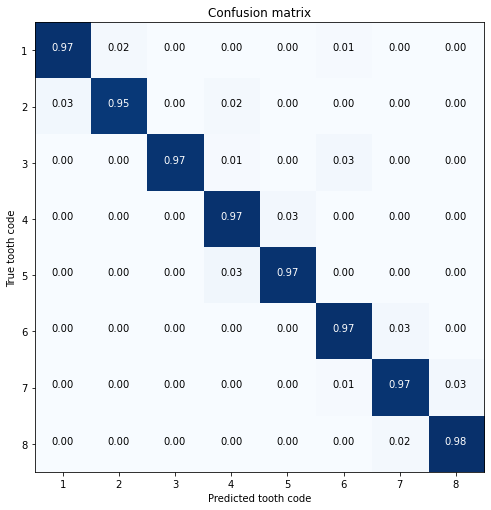}}
	\caption{Confusion matrix for tooth identification. (a) Result of the direct method. (b) Result of the proposed method.}\label{fig:cf_mat}
\end{figure}

This subsection presents the performance evaluation of tooth identification. The precision, recall, and F1-score were used to evaluate the results of the identification method. In Step 2-1, teeth are initially classified into four types, instead of directly predicting the eight tooth codes. The direct identification method can often misclassify teeth within the same tooth type.  As shown in Fig. \ref{fig:cf_mat}a, the direct method confuses first premolars (code 4) and wisdom teeth (code 8) in particular. These errors hinder the performance improvement of the direct method. However, the four type-based method achieves a high accuracy by preventing misclassification due to similar tooth shape. Table \ref{tbl:comparison_id} and Fig. \ref{fig:cf_mat} show that the proposed method leads to more accurate identification.

\begin{table}[!h]
	\footnotesize
	\centering
	\caption{Quantitative evaluation for tooth identification methods. \label{tbl:comparison_id}}\vskip 0.0in
	\begin{tabular}{ccc} \hline
		{Metric} & {\bf Direct method} & {\bf Proposed method} \\ \cmidrule(lr){1-1} \cmidrule(lr){2-2} \cmidrule(lr){3-3}
		{Precision ($\%$)} & {$93.05 \pm 7.85$} & {$96.81 \pm 1.67$} \\
		{Recall ($\%$)} & {$87.22 \pm 7.42$} & {$90.13 \pm 5.30$} \\
		{F1-score ($\%$)} & {$90.04 \pm 7.63$} & {$93.35 \pm 2.54$} \\ \hline
	\end{tabular}
\end{table}

\subsubsection{Individual tooth segmentation}

To evaluate the 2D and 3D tooth segmentation performances, we used precision, recall, Dice similarity coefficient (DSC) \cite{raudaschl2017}, Hausdorff distance (HD) \cite{raudaschl2017}, and average symmetric surface distance (ASSD) \cite{raudaschl2017}.

\textit{2D individual tooth segmentation.} The proposed method proceeds in two steps consisting of the bounding box detection in Step 2-1 and individual segmentation in Step 2-3, while Mask R-CNN \cite{he2017} achieves the same task in a single step. We implemented both approaches, and reported the quantitative evaluations in Table \ref{tbl:comparison_maskrcnn}. The proposed method is numerically more accurate than the Mask R-CNN.

The accuracy of 2D segmentation is important because the key of precise 3D tooth segmentation is to use tight tooth ROIs obtained by 2D segmentation. Although the detection and segmentation are not performed simultaneously, two simple convolutional networks (one-stage object detection and U-shaped FCN) are designed to achieve a high accuracy.

\begin{table}[!h]
	\footnotesize
	\centering
	\caption{Quantitative evaluation for 2D tooth segmentation methods.\label{tbl:comparison_maskrcnn}}\vskip 0.0in
	\begin{tabular}{ccc} \hline
		{Metric} & {\bf 2DMask R-CNN} & {\bf Proposed method} \\ \cmidrule(lr){1-1} \cmidrule(lr){2-2} \cmidrule(lr){3-3}
		{Precision ($\%$)} & {$92.17 \pm 5.55$} & {$96.18 \pm 4.67$} \\
		{Recall ($\%$)} & {$90.35 \pm 5.87$} & {$93.17 \pm 5.83$} \\
		{DSC ($\%$)} & {$90.98 \pm 3.25$} & {$94.41 \pm 3.10$} \\
		{HD ($mm$)} & {$1.64 \pm 0.93$} & {$1.34 \pm 0.86$} \\
		{ASSD ($mm$)} & {$0.39 \pm 0.12$} & {$0.27 \pm 0.12$} \\ \hline
	\end{tabular}
\end{table}

\begin{table*}[!h]
	\scriptsize
	\centering
	\caption{Quantitative comparison for 3D tooth segmentation methods.}\label{tbl:comparison_seg}\vskip 0.0in
	\begin{tabular}{cccccccccc} \hline
		& \multicolumn{2}{c}{\bf Mask R-CNN} & \multicolumn{2}{c}{\bf PANet} & \multicolumn{2}{c}{\bf HTC} & \multicolumn{2}{c}{\bf ToothNet} & {\bf Loose \& tight ROIs}\\ 
		{Metric} & {mean $\pm$ std} & {$p$-value} & {mean $\pm$ std} & {$p$-value} & {mean $\pm$ std} & {$p$-value} & {mean $\pm$ std} & {$p$-value} & {mean $\pm$ std}\\ \cmidrule(lr){1-1} \cmidrule(lr){2-3} \cmidrule(lr){4-5} \cmidrule(lr){6-7} \cmidrule(lr){8-9} \cmidrule(lr){10-10}
					 
		{Precision ($\%$)} & {$93.98 \pm 11.07$} & {$<0.001$} & {$93.61 \pm 6.88$} & {$<0.001$} & {$91.43 \pm 4.28$} & {$<0.001$} & {$89.40 \pm 5.84$} & {$<0.001$} & {$95.97 \pm 2.00$}\\
		{Recall ($\%$)} & {$88.82 \pm 10.69$} & {$<0.01$} & {$89.76 \pm 7.89$} & {$<0.001$} & {$92.31 \pm 6.23$} & {$<0.001$} & {$93.25 \pm 5.71$} & {$<0.001$} & {$93.71 \pm 2.08$} \\
		{DSC ($\%$)} & {$90.75 \pm 10.48$} & {$<0.001$} & {$91.04 \pm 6.32$} & {$<0.001$} & {$91.59 \pm 4.57$} & {$<0.001$} & {$91.40 \pm 5.07$}& {$<0.001$} & {$94.79 \pm 1.34$} \\
		{HD ($mm$)} & {$3.09 \pm 2.01$} & {$<0.001$} & {$2.84 \pm 1.88$} & {$<0.001$} & {$2.60 \pm 1.79$} & {$<0.001$} & {$2.53 \pm 1.97$} & {$<0.001$} & {$1.66 \pm 0.72$} \\
		{ASSD ($mm$)} & {$0.36 \pm 0.31$} & {$<0.001$} & {$0.32 \pm 0.26$} & {$<0.001$} & {$0.26 \pm 0.21$} & {$<0.001$} & {$0.27 \pm 0.20$} & {$<0.001$} & {$0.14 \pm 0.04$} \\ \hline
	\end{tabular}
\end{table*}

\begin{table}[!h]
	\scriptsize
	\centering
	\caption{Ablation study for the proposed method.}\label{tbl:abl_study}\vskip 0.0in
	\begin{tabular}{ccccc} \hline
		& \multicolumn{2}{c} {\bf Loose ROI} & \multicolumn{2}{c}{\bf Tight ROI}\\ 
		{Metric} & {mean $\pm$ std} & {$p$-value} & {mean $\pm$ std} & {$p$-value}\\ 
		\cmidrule(lr){1-1} \cmidrule(lr){2-3} \cmidrule(lr){4-5} 
					 
		{Precision ($\%$)} & {$94.80 \pm 6.53$} & {$<0.001$} & {$94.70 \pm 2.79$} & {$<0.001$}\\
		{Recall ($\%$)} & {$92.98 \pm 3.68$} & {$<0.001$} & {$91.54 \pm 2.12$} & {$<0.001$}\\
		{DSC ($\%$)} & {$93.76 \pm 4.81$} & {$<0.001$} & {$92.98 \pm 1.75$} & {$<0.001$}\\
		{HD ($mm$)} & {$2.37 \pm 1.93$} & {$<0.001$} & {$1.98 \pm 1.20$} & {$<0.001$}\\
		{ASSD ($mm$)} & {$0.20 \pm 0.20$} & {$<0.001$} & {$0.19 \pm 0.05$} & {$<0.001$}\\ \hline
	\end{tabular}
\end{table}

\textit{3D individual tooth segmentation.} We developed a fully automated multi-step method for 3D individual segmentation. To verify the effectiveness of the proposed method, we compared it with Mask R-CNN \cite{he2017}, Path aggregation network (PANet) \cite{liu2018}, Hybrid task cascade (HTC) \cite{chen2019hybrid}, and ToothNet \cite{cui2019}. These methods cannot be applied to large size 3D CBCT images directly because of the computational limit. For comparison experiments, we implemented the methods in a patch-based fashion as an alternative to avoid the limitation. We also performed an ablation study of the proposed method by using either loose or tight ROI, or both. The four methods adopted for comparison show lower performances, as shown in Table \ref{tbl:comparison_seg}. These methods perform individual segmentation from the original CBCT images. In contrast, the proposed method has the advantage of using loose and tight ROIs that provide considerable background region in advance. In particular, the tight ROI excludes structures on sides (\textit{e.g.}, adjacent teeth, jaw, etc) of the target tooth. When using only tight ROIs, the recall is the lowest because loss of tooth information may occur where the tight ROI boundary intersects the tooth boundary, as shown in Table \ref{tbl:abl_study}. The use of only loose ROI shows that HD tends to be high because there is no information on tooth boundaries. A combination of the two ROIs enhances the segmentation performance, as the tight ROI provides detailed information on the target tooth and the loose ROI compensates for the disadvantage of the tight ROI. Wilcoxon signed-rank test \cite{wilcoxon1992} was used to calculate the statistical significance differences between the proposed method and other methods, as summarized in Tables \ref{tbl:comparison_seg} and \ref{tbl:abl_study}.

\subsection{Metal artifact-contaminated CBCT images}
\begin{figure}[!h]
\centering
\includegraphics[width=0.48\textwidth]{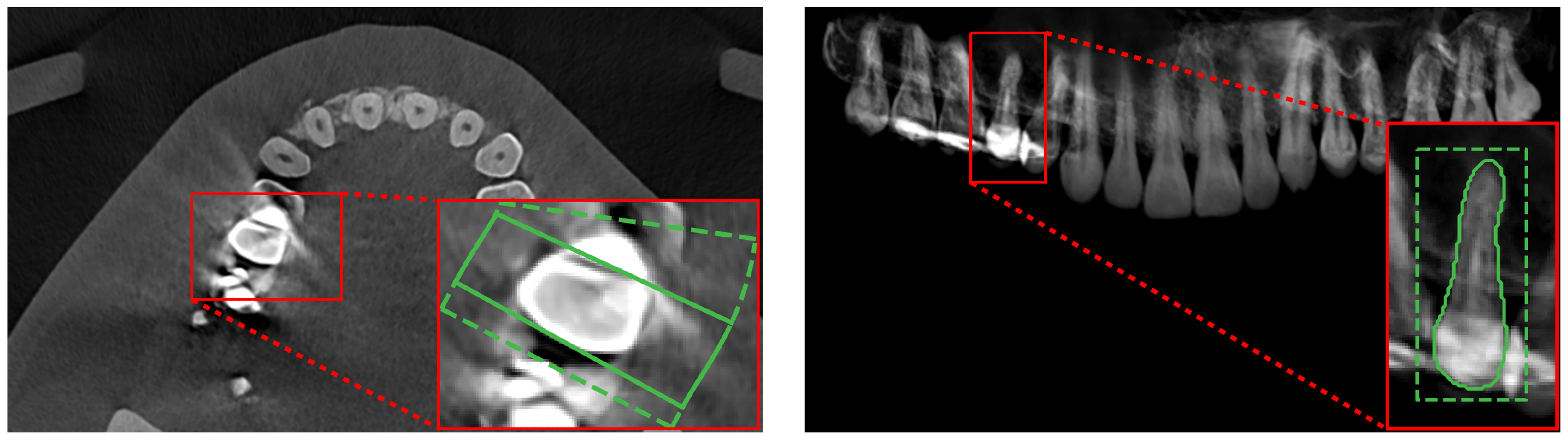}
\caption{Image on the left is a CBCT image that is affected by metal artifacts. Image on the right is a panoramic image generated from Step 1, which is not affected by metal-related artifacts.}
\label{fig:metal}
\end{figure}

\begin{figure*}[h!]
  	\centering
	\subfigure[] {\includegraphics[width=0.13\textwidth]{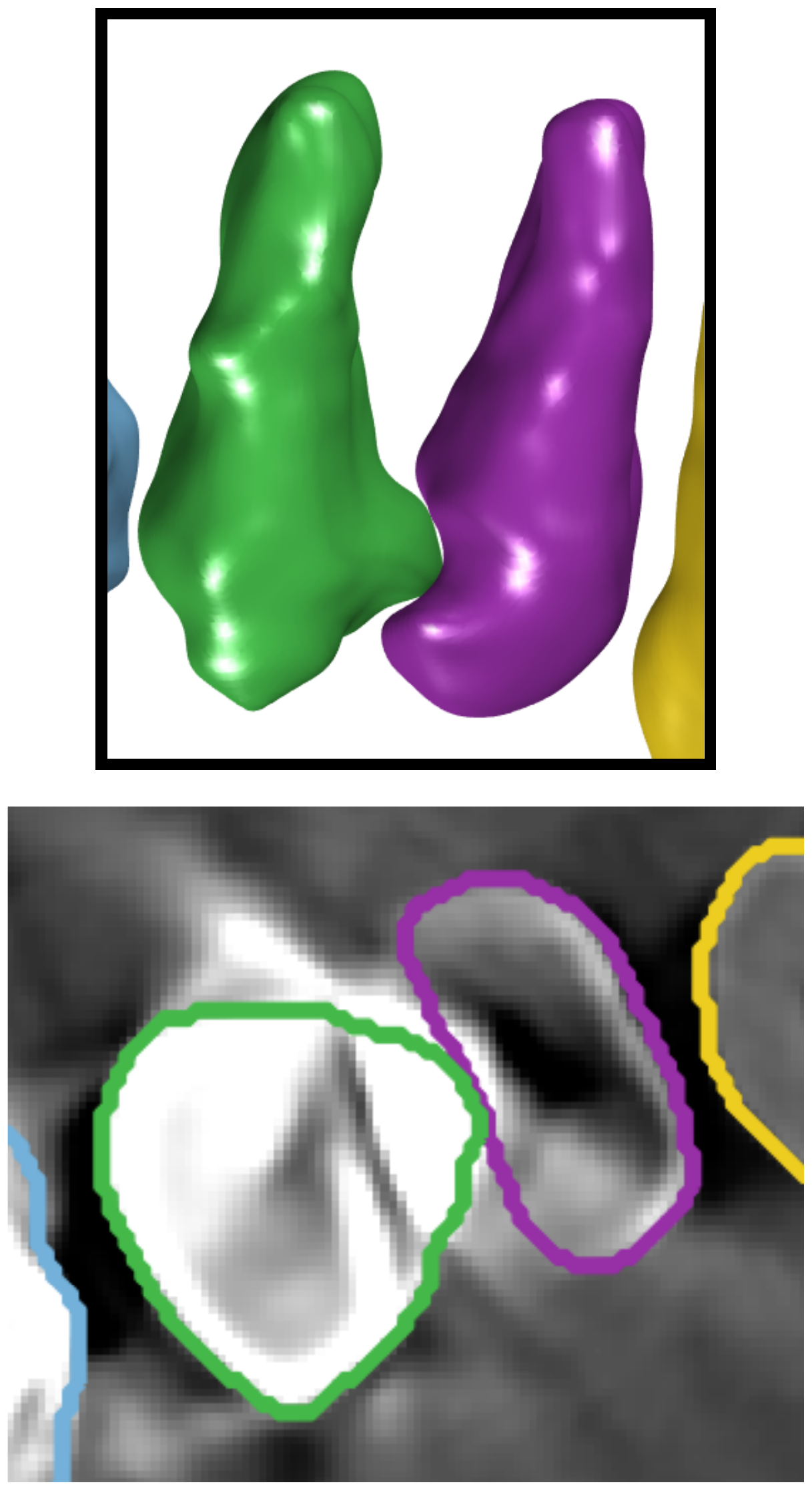}}~
	\subfigure[] {\includegraphics[width=0.13\textwidth]{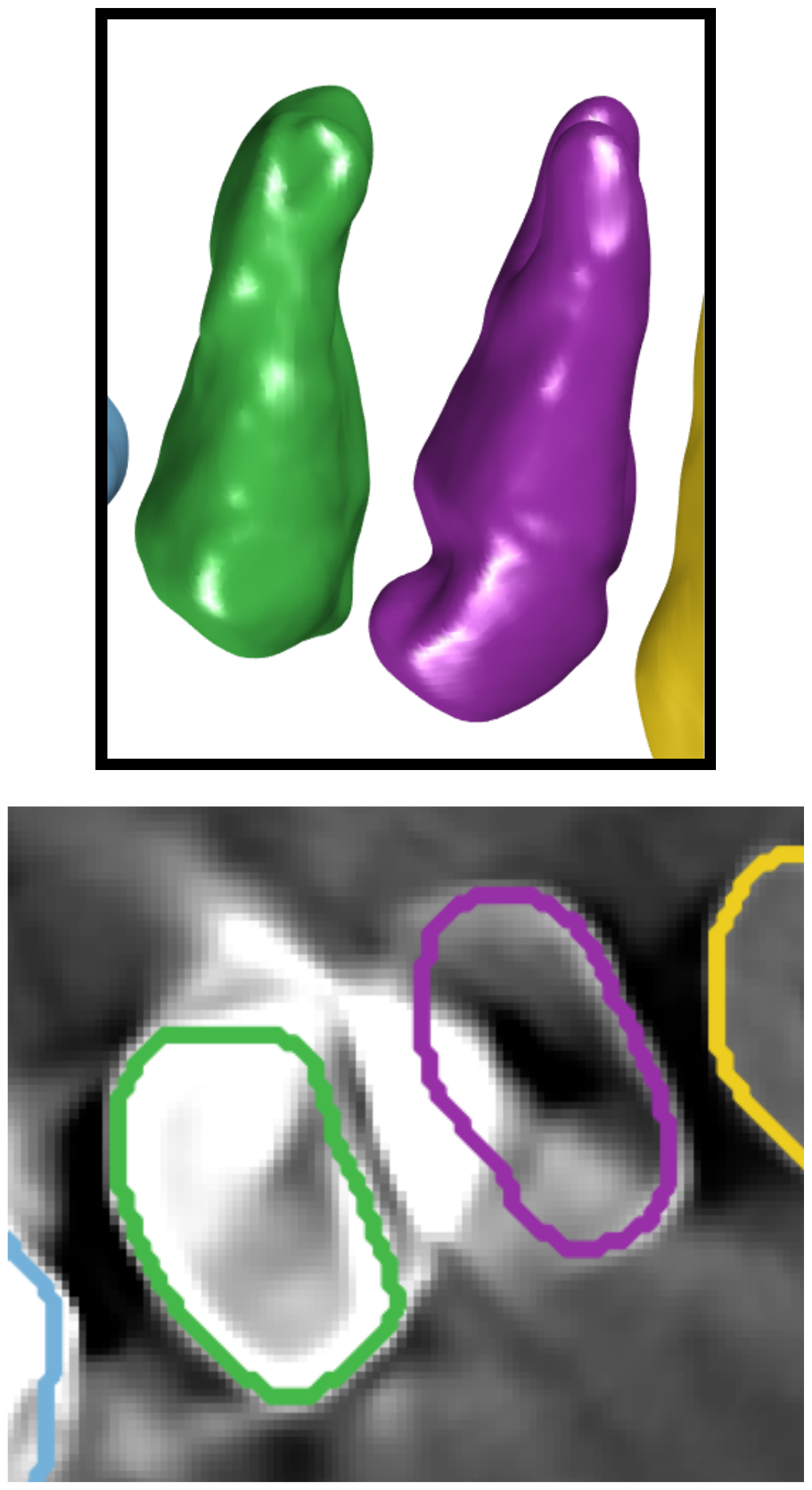}}~
	\subfigure[] {\includegraphics[width=0.13\textwidth]{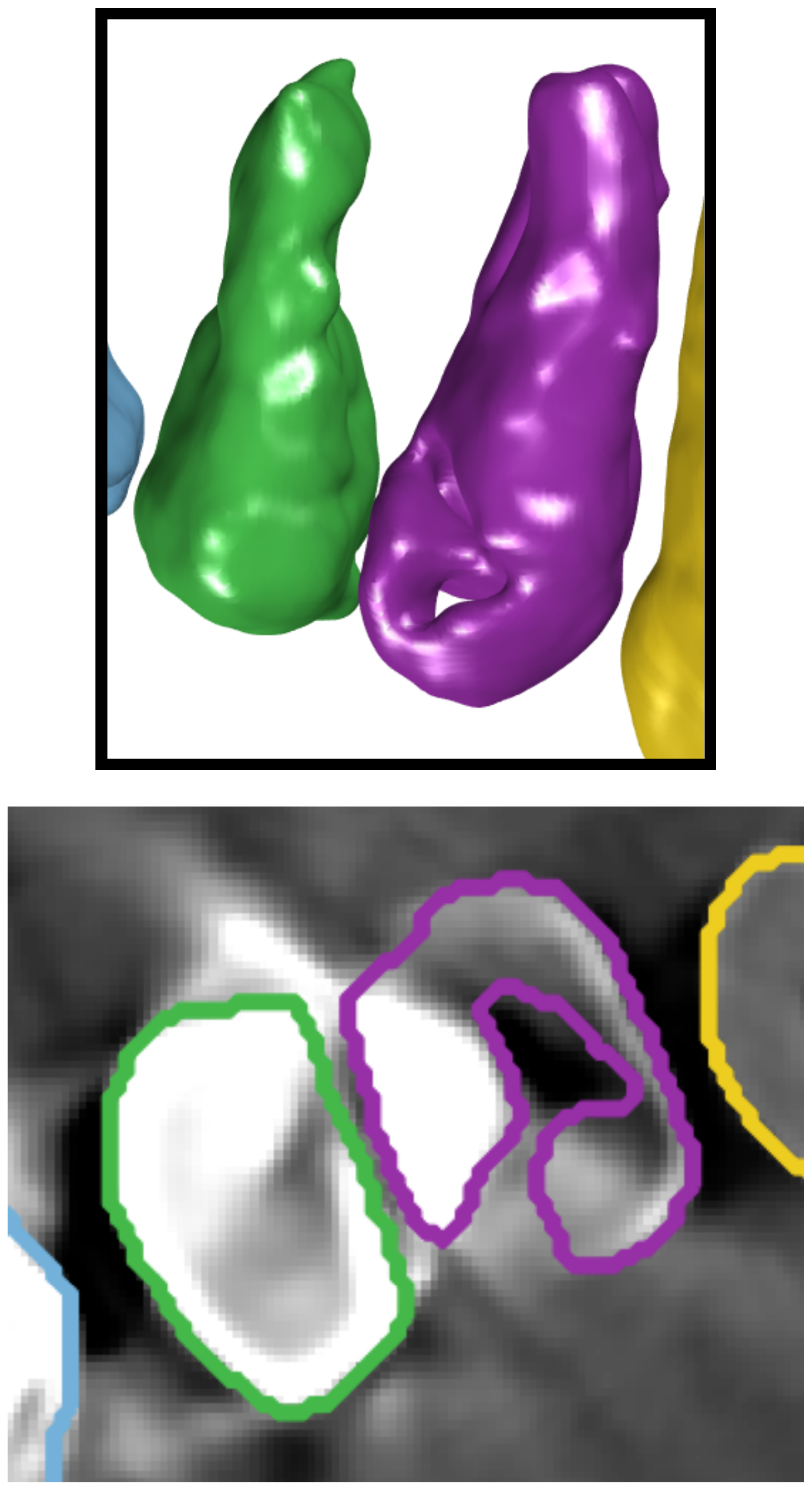}}~
	\subfigure[] {\includegraphics[width=0.13\textwidth]{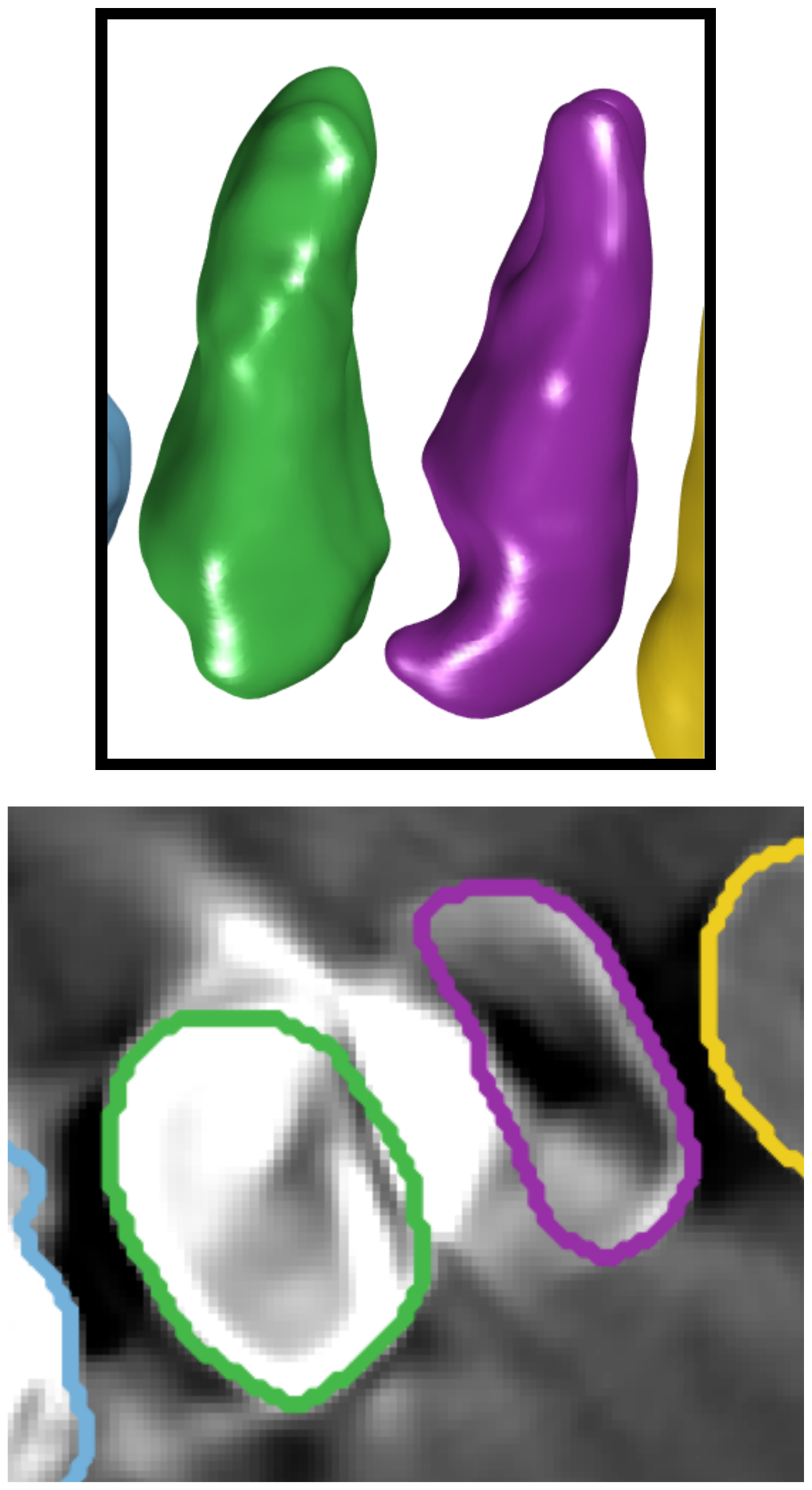}}~
	\subfigure[] {\includegraphics[width=0.13\textwidth]{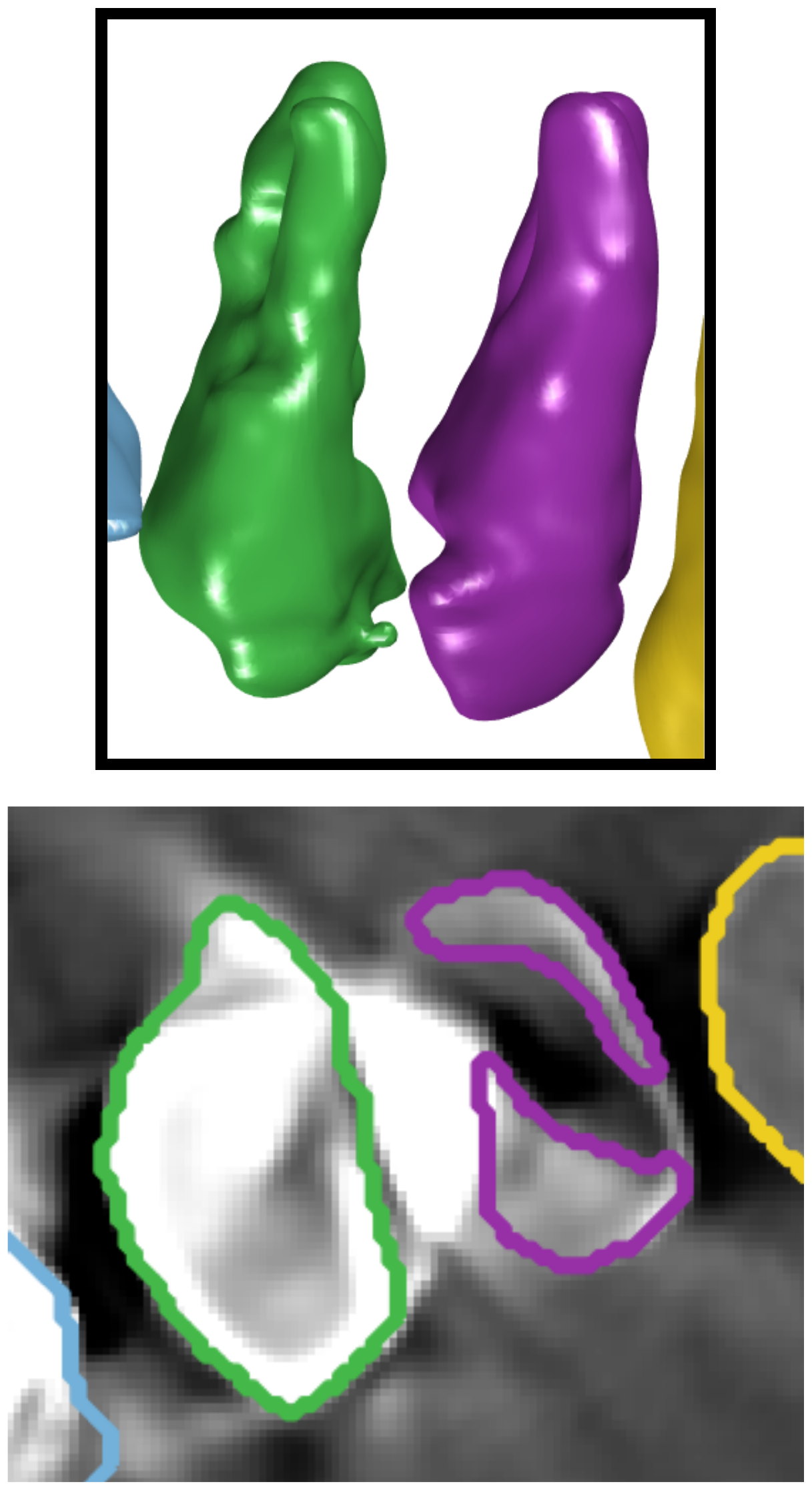}}~
	\subfigure[] {\includegraphics[width=0.13\textwidth]{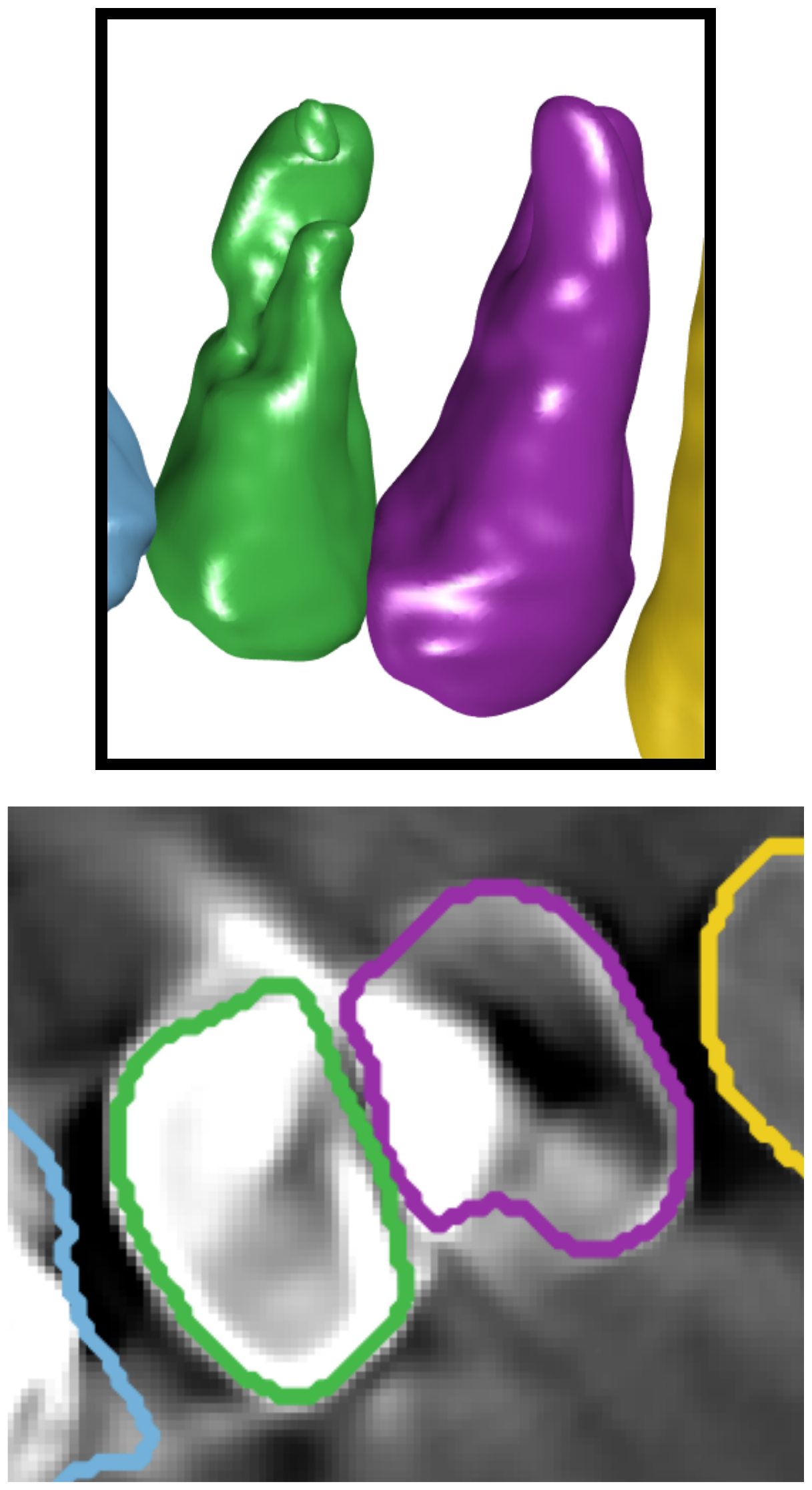}}~
	\subfigure[] {\includegraphics[width=0.13\textwidth]{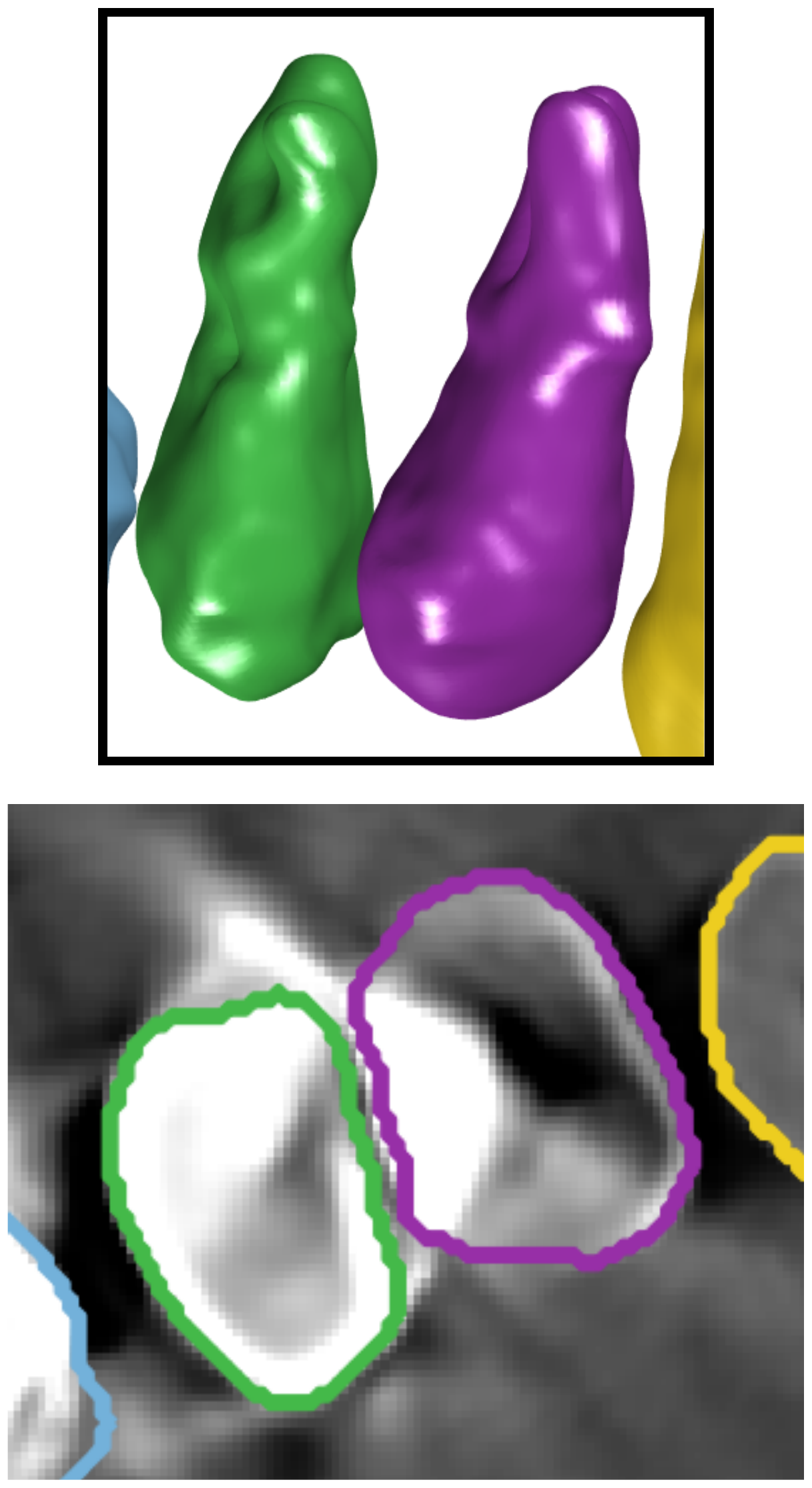}}

  	\caption{Qualitative comparison for 3D individual tooth segmentation in a CBCT image with metal artifacts. Segmentation result of (a) Mask R-CNN, (b) PANet, (c) HTC, (d) ToothNet, (e) the proposed method using loose ROIs, (f) tight ROIs, and (g) both loose and tight ROIs.}\label{fig:comparison_metal}
\end{figure*}

The proposed method effectively handles problems that are caused by metal-related artifacts. Fig. \ref{fig:metal} shows that the CBCT image is severely contaminated by metal artifacts, whereas metal artifacts are significantly reduced in the corresponding panoramic image generated by the CBCT image. The panoramic images in Step 1 allow to accurately perform 2D tooth detection and segmentation. These results provide prior knowledge of each 3D tooth as loose and tight ROIs. As shown in Fig. \ref{fig:metal}, the tight ROI excludes adjacent teeth, even though the tooth boundaries are obscured by metal artifacts. Fig. \ref{fig:comparison_metal} shows a qualitative evaluation for 3D tooth segmentation in a CBCT image with metal artifacts. As shown in Figs. \ref{fig:comparison_metal}f and \ref{fig:comparison_metal}g, the segmentation results in the degraded CT image are superior to those of Figs. \ref{fig:comparison_metal}a, \ref{fig:comparison_metal}b, \ref{fig:comparison_metal}c, \ref{fig:comparison_metal}d, and \ref{fig:comparison_metal}e as the tight ROIs provide robust tooth boundary information. However, Fig. \ref{fig:comparison_metal}f presents that using only the tight ROIs may not provide precise segmentation because it can cut out the edges of the teeth. Therefore, Fig. \ref{fig:comparison_metal}g illustrates the advantages and effectiveness of the proposed method using both loose and tight ROIs.

\subsection{Missing tooth}
\begin{figure}[!h]
\centering
\includegraphics[width=0.36\textwidth]{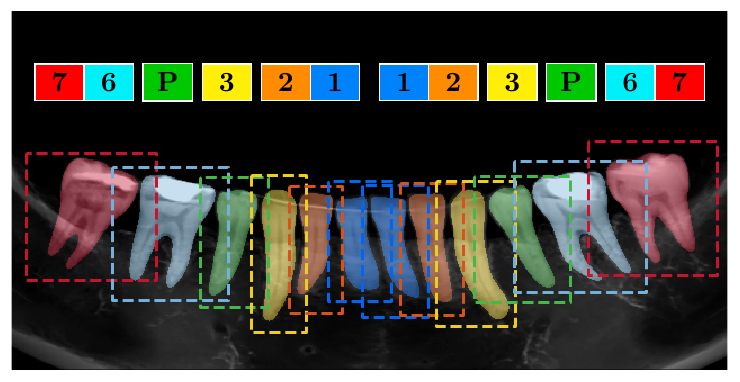}
\caption{Illustration of tooth identification when there are missing teeth. Two premolar (class P) corresponding to number 4 are missing.}
\label{fig:discussion2}
\end{figure}

People often lose their teeth due to factors such as cavities, periodontal disease, aging, dental trauma and orthodontic treatment. To address these cases, we applied CBCT images with missing teeth (except for wisdom teeth) to the proposed method. Four type-based classification was successful, but tooth identification was incomplete. With exception of canines, each tooth quadrant contains two or more teeth of the same type. When numbering the neighboring tooth of the same type as a missing tooth, it is difficult to identify the neighboring tooth in panoramic images. Therefore, we suggest performing only classification in the case of a missing tooth, as shown in Fig. \ref{fig:discussion2}.

\section{Discussion and Conclusion}
\label{sec:discussion}

In this paper, we developed a fully automated segmentation and identification method for individual teeth and jaws from CBCT images. Given CBCT data, the method automatically generates the maxillary and mandibular panoramic images that are projected along the reference curve representing a region-based shape feature of a dental arch. In the maxillary and mandibular panoramic images, 2D tooth segmentation and identification are performed using deep learning methods, which are vital in high-precision 3D tooth segmentation and identification. Experiments showed that the accuracy of the method is suitable for the clinical setting. Our method overcomes the limitations of existing automated methods by achieving fully automation and improved accuracy. Additionally, the method addresses the difficulty of learning high-dimensional data.

The main idea of the proposed method is the careful use of the accurate and robust 2D tooth segmentation and identification in 2D panoramic images in an indirect manner to address the difficulty of 3D segmentation from metal artifact-contaminated 3D CBCT images. In a clinical dental CBCT environment (\textit{e.g.}, low dose radiation exposure), metal related-artifacts are common. The proposed method utilizes the crucial observation that metal artifacts are significantly reduced in the upper and lower panoramic images generated from the CBCT images. The outcome in Step 2 serves as strong prior knowledge of 3D tooth segmentation, which plays an important role in separating teeth from 3D images, in cases where teeth are often contacted, overlapped, or connected owing to metal-related artifacts.

The automated system proposed in this study improves the efficiency of dentists by reducing the cumbersome and time-consuming manual intervention. The result provides an improved workflow for dentists to simulate pre-operative orthodontic treatment and manufacture implant surgical guides. Digital occlusion analysis is potentially possible by combining our method with the intra-oral scan model \cite{xu2018,tian2019} via registration. Hence, it is expected to play an important role in digital dentistry.

\section*{Acknowledgements}

This research was supported by a grant of the Korea Health Technology R$\&$D Project through the Korea Health Industry Development Institute (KHIDI), funded by the Ministry of Health $\&$ Welfare, Republic of Korea (HI20C0127). We would like to express our deepest gratitude HDXWILL which shares dental CBCT images and ground-truth data.

\bibliographystyle{IEEEtran}
\bibliography{manuscript.bib}

\end{document}